\documentclass[letterpaper, 10 pt, conference]{ieeeconf}  

\IEEEoverridecommandlockouts                              

\usepackage{graphicx} 
\usepackage{amsmath} 

\usepackage{subcaption}
\usepackage{algorithmicx,algpseudocode}

\algnewcommand\algorithmicswitch{\textbf{switch}}
\algnewcommand\algorithmiccase{\textbf{case}}
\algnewcommand\algorithmicassert{\texttt{assert}}
\algnewcommand\Assert[1]{\State \algorithmicassert(#1)}%
\algdef{SE}[SWITCH]{Switch}{EndSwitch}[1]{\algorithmicswitch\ #1\ \algorithmicdo}{\algorithmicend\ \algorithmicswitch}%
\algdef{SE}[CASE]{Case}{EndCase}[1]{\algorithmiccase\ #1}{\algorithmicend\ \algorithmiccase}%
\algtext*{EndSwitch}%
\algtext*{EndCase}%

\title{\LARGE \bf
Towards CNN Map Compression for camera relocalisation 
}

\author{Luis Contreras$^{1}$ and Walterio Mayol-Cuevas$^{2}$
\thanks{*This work was partially supported by CONACYT and the Secretaria de Educacion Publica, Mexico}%
\thanks{Department of Computer Science, University of Bristol, United Kingdom}%
\thanks{$^{1}${\tt\small cslact@bristol.ac.uk}}%
\thanks{$^{2}${\tt\small wmayol@cs.bris.ac.uk}}%
}

\begin{document}

\maketitle
\thispagestyle{empty}
\pagestyle{empty}

\begin{abstract}

This paper presents a study on the use of Convolutional Neural Networks for camera relocalisation and its application to map compression. We follow state of the art visual relocalisation results and evaluate response to different data inputs -- namely, depth, grayscale, RGB, spatial position and combinations of these. We use a CNN map representation and introduce the notion of CNN map compression by using a smaller CNN architecture. We evaluate our proposal in a series of publicly available datasets. This formulation allows us to improve relocalisation accuracy by increasing the number of training trajectories while maintaining a constant-size CNN.

\end{abstract}

\section{INTRODUCTION}
\label{sec:introduction}

Following our work on point cloud compression mapping via feature filtering in \cite{contreras:2015} and \cite{contreras:2017}, we aim to generate compact map representations useful for camera relocalisation via compact CNNs. To ask what is the minimal map representation that enables later use is a meaningful question that underpins many applications for moving agents.

In this work, we specifically explore a neural network architecture tested for the relocalisation task. We study the response of such architecture to different inputs -- e.g. color and depth images --, and the relocalisation performance of pre-trained neural networks in different tasks. 

Biologically inspired visual models have been proposed for a while \cite{harvey:1991}, \cite{milford:2004}. How humans improve learning after multiple training of the same view and how they filter useful information have also been an active field of study. One widely accepted theory of the human visual system suggests that a number of brain layers sequentially interact from the signal stimulus to the abstract concept \cite{dicarlo:2012}. Under this paradigm, the first layers -- connected directly to the input signal -- are a series of specialized filters that extract very specific features, while deeper layers infer more complex information by combining these features.
%
%
%
%

Finally, overfitting a neural network by excessive training with the same dataset is a well known issue; rather, here we study how the accuracy improves by revisiting the same area several times introducing new views to the dataset.

This paper is organized as follows. In Section \ref{sec:relatedwork} we discuss work related to convolutional neural networks and camera pose. Our CNN for relocalisation is then introduced in Section \ref{sect:cnn}, where we describe its architecture. Then, Section \ref{sect:cnnmap} introduces the notion of CNN map representation and compression. Experimental results are presented in Section \ref{sect:experiments}; finally, we  outline our  discussion and conclusions.

\section{RELATED WORK}
\label{sec:relatedwork}

Even though neural networks are not a novel concept, due to the increase in computational power, their popularity has grown in recent years \cite{bengio:courville:2012} \cite{bengio:courville:2013}. 

Related to map compression, dimensionality reduction through neural networks was first discussed in \cite{hinton:2006}. In \cite{chatfield:simonyan:2011} an evaluation to up-to-date data encoding algorithms for object recognition was presented, and it was extended in \cite{chatfield:simonyan:2014} to introduce the use of Convolutional Neural Networks for the same task.

\cite{agrawal:2015} introduced the idea of egomotion in CNN training by concatenating the output of two parallel neural networks with two different views of the same image; at the end, this architecture learns valuable features independent of the point of view.

In \cite{jarrett:2009}, Jarrett et al. concluded that sophisticated architectures compensate for lack of training. Garg et al. \cite{garg:2016} explore this idea for single view depth estimation where they present a stereopsis based auto-encoder that uses few instances on the KITTI dataset. Then, \cite{eigen:2014}, \cite{li:shen:2015}, and \cite{liu:shen:2015} continued studying the use of elaborated CNN architectures for depth estimation.

Moving from depth to pose estimation was a logical step. One of the first 6D camera pose regressors was presented in \cite{Gee:Mayol:2012} via a general regression NN (GRNN) with synthetic poses. More recently, PoseNet is presented in \cite{kendall:grimes:2015}, where they regress the camera pose using a CNN model. This idea is also explored in \cite{long:kneip:2016} for image matching via training a CNN for frame interpolation through video sequences.



\section{The Relocalisation CNN} \label{sect:cnn}

\begin{figure}[h]
	\begin{center}
	\includegraphics[width=0.45\textwidth]{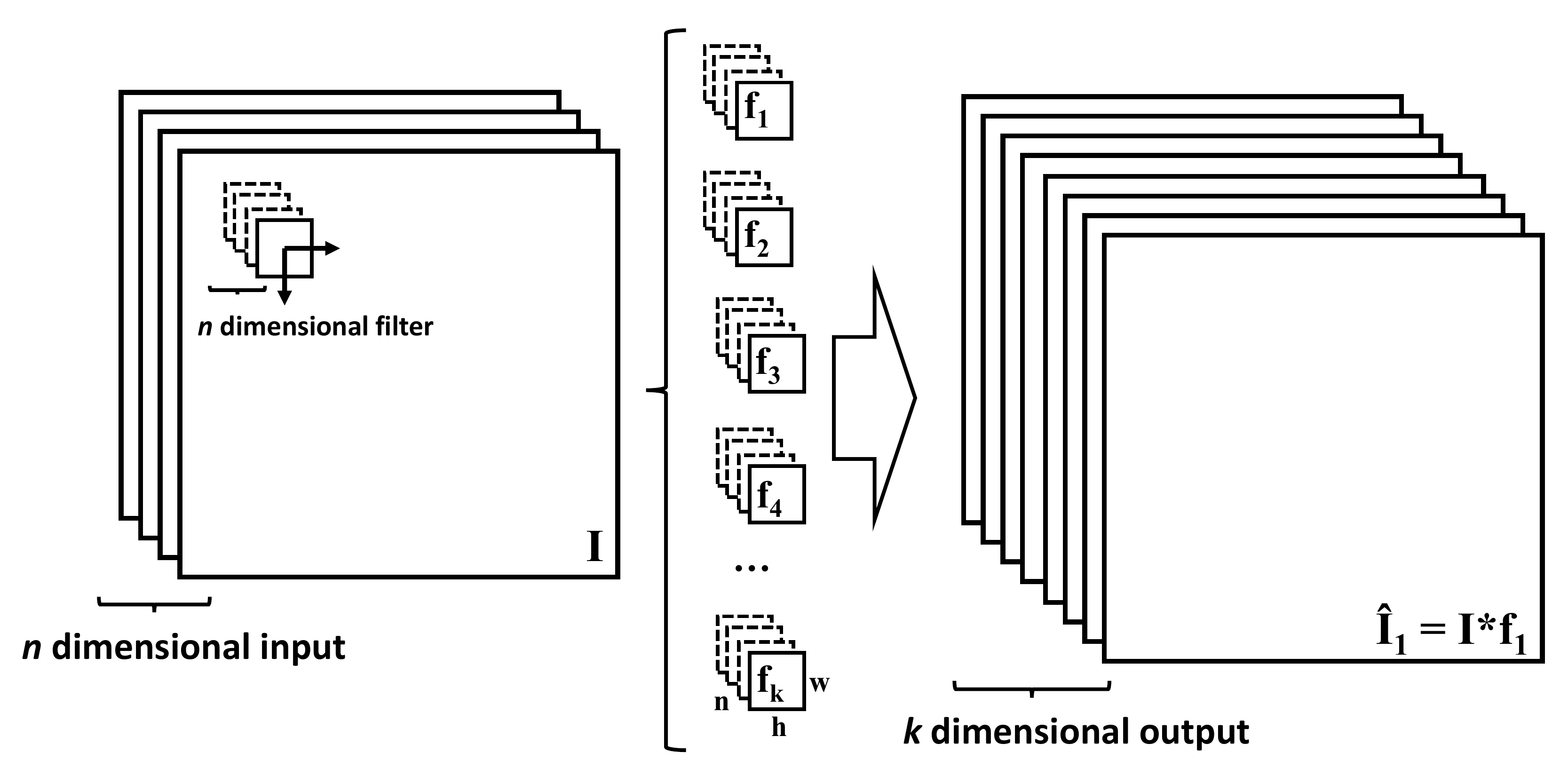}
	\caption{Convolutional Neural Network (CNN) elements. It consist of an input $I$, a series filters $f_k$, and its associated output $\hat{I}_k$. The filter depth depend on the number of input channels.}
	\label{fig:cnnstruct}
	\end{center}
\end{figure}

We develop a Convolutional Neural Network, or CNN, to address the camera relocalization problem. A CNN can be considered as a filter bank where the filters' weights are such as they minimize the error between an expected output and the system response to a given input. In Figure \ref{fig:cnnstruct} we show the elements from one layer to the next in a typical CNN architecture -- a more detailed CNN implementation can be found in specialized works such as \cite{srinivas:2016} and \cite{nielsen:2015}. For a given input $I$ and a series of $k$ filters $f_k$, it is generated an output $\hat{I}_k = I*f_k$; the filters $f_k$ can be initialized randomly or pre-trained in a different task.

\begin{figure}[h]
	\begin{subfigure}{.15\textwidth}
		\centering
		\includegraphics[width=1\textwidth]{{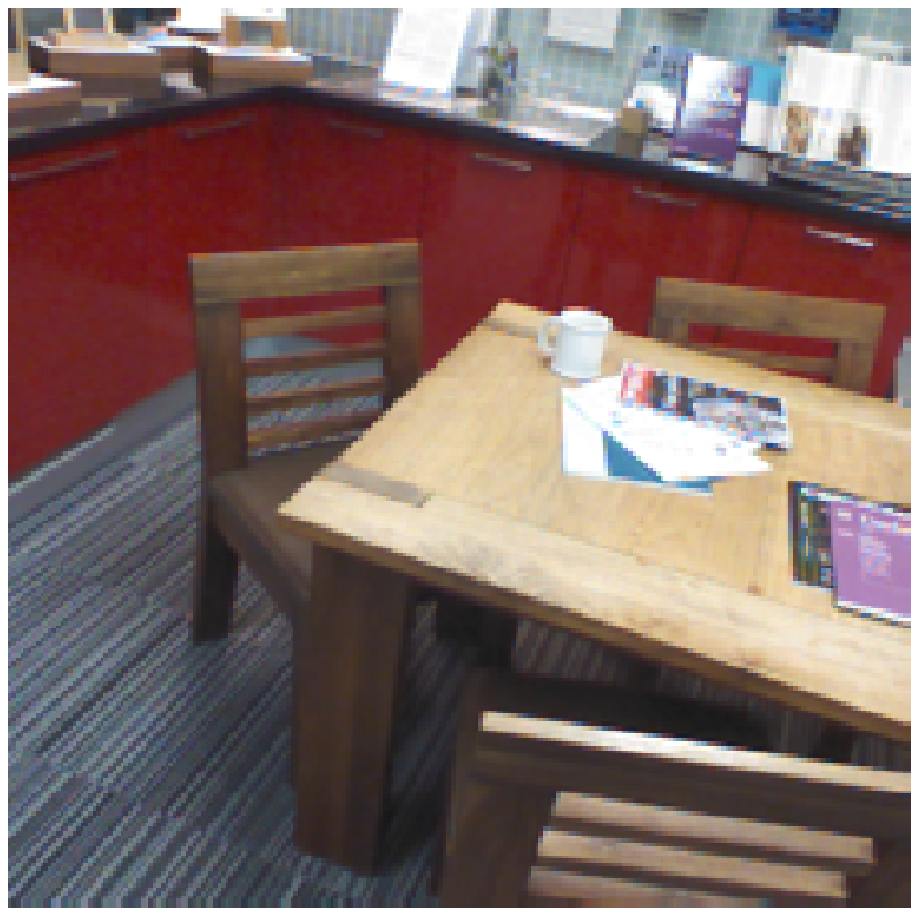}}
		\caption{}
		\label{fig:datargb}
	\end{subfigure}
	\begin{subfigure}{.15\textwidth}
  		\centering
  		\includegraphics[width=1\textwidth]{{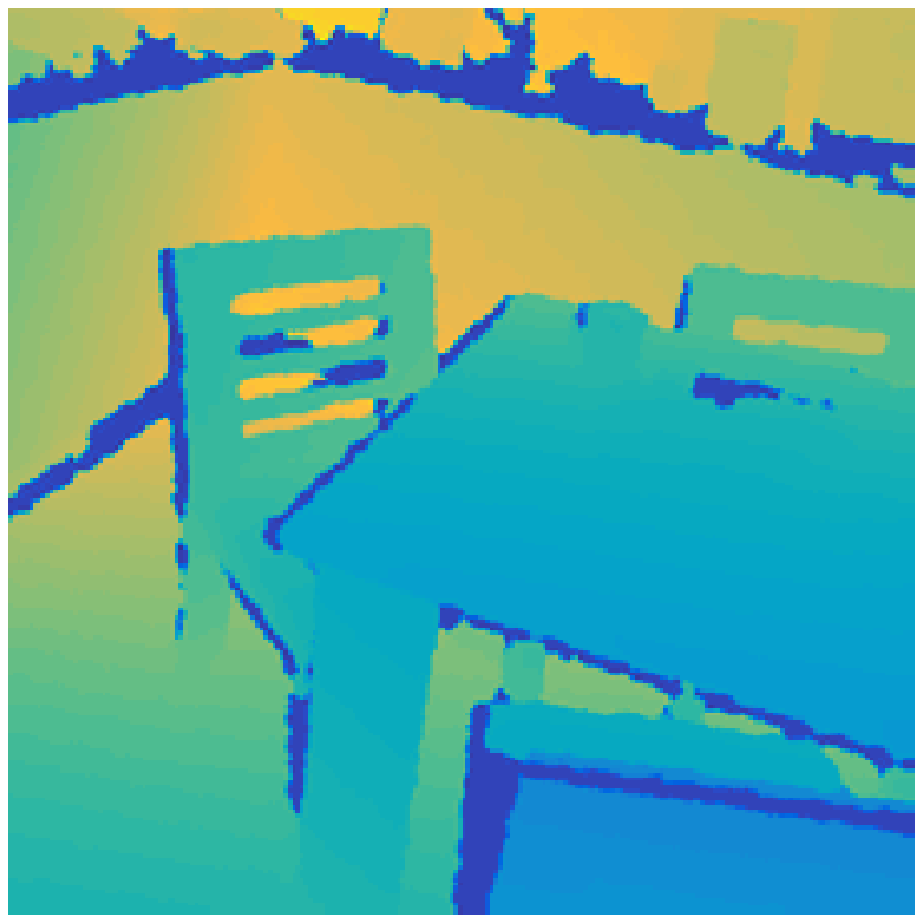}}
  		\caption{}
  		\label{fig:datadepth}
	\end{subfigure}
	\begin{subfigure}{.15\textwidth}
  		\centering
  		\includegraphics[width=1\textwidth]{{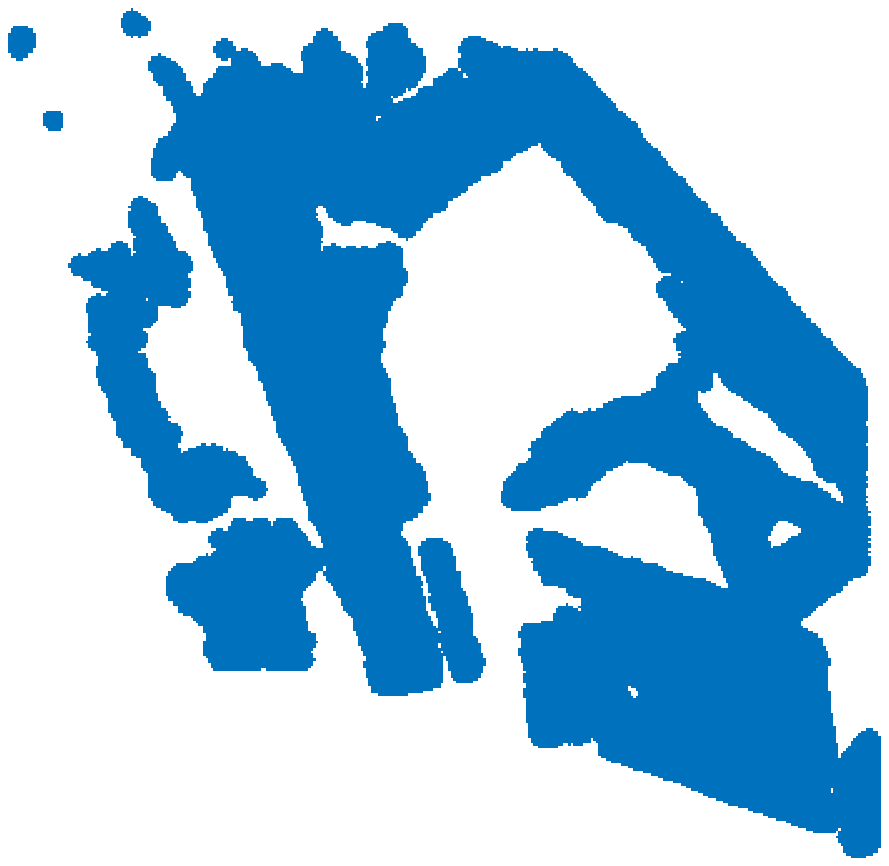}}
  		\caption{}
  		\label{fig:datapc}Relocalisation 
	\end{subfigure}
	\caption{Different data input samples: a) color image, b) depth map, and c) 3D point cloud.}
	\label{fig:dataall}
\end{figure}

We highlight the direct relationship between the input channels and the filters' depth, because we will work with different n-dimensional inputs. We evaluate the performance in a number of dataset both RGB and RGBD -- more specifically, we tested tensors with the following information per element: gray and RGB values, depth distance, 3D spatial position, RGB+depth, and RGB+spatial position; some typical inputs are shown in Figure \ref{fig:dataall}. The original input is pre-processed by cropping the central area and resizing it, generating 224x224 arrays.

Our study is based in the PoseNet model as described in \cite{kendall:grimes:2015}. The expected output is a 6 DOF pose $p = [x, q]$, where $x$ is the spatial position and $q$ is the orientation in quaternion form. We use PoseNet loss function:

$$loss(I)=\Vert\hat{x}-x\Vert_2+\beta\left\Vert\hat{q}-\frac{q}{\Vert q \Vert}\right\Vert_2$$

For ease, instead of the GoogLeNet arquitecture \cite{szegedy:2014}, we use a fast convolutional neural network (CNN-F) implementation as in \cite{chatfield:simonyan:2014} -- an architecture with 8 layers: 5 convolutional and three fully-connected. We introduce a couple of changes, as follows: the dimension in the first layer depends on the input \textit{n}; in addition, the final fully-connected layer size changes to the pose vector length. Table \ref{tab:cnnf} details our network.

\begin{table}[!h]
	\caption{Fast Convolutional Neural Netwirk (CNN-F) architecture, as in \cite{chatfield:simonyan:2014}. The dimension \textit{n} in \textit{conv1} depends on the input data; the output in \textit{full8} is the pose vector.}
	\label{tab:cnnf}
	\begin{center}
	\resizebox{0.48\textwidth}{!}{
		\begin{tabular}{|c|c|c|c|}
		\hline
			conv1&conv2&conv3&conv4\\ 
		\hline
			11x11xnx64&5x5x64x256&3x3x256x256&3x3x256x256 \\
		\hline
			conv5&full6&full7&full8 \\
		\hline
			3x3x256x256&4096&4096&7 \\
		\hline
		\end{tabular}}
	\end{center}
\end{table}

\begin{figure}[h]
	\begin{subfigure}{.1125\textwidth}
		\centering
		\includegraphics[bb = 0 0 640 360, width=1\textwidth]{{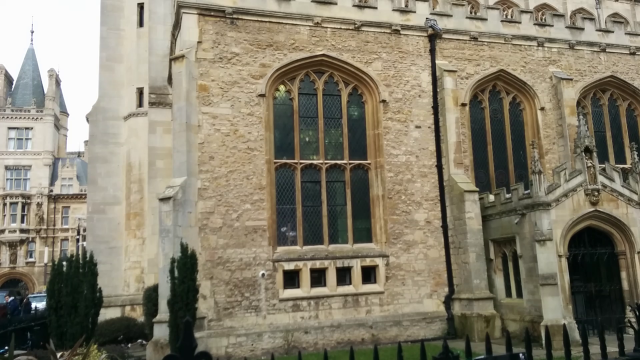}}
		\caption{}
		\label{fig:posenet1}
	\end{subfigure}
	\begin{subfigure}{.1125\textwidth}
  		\centering
  		\includegraphics[bb = 0 0 640 360, width=1\textwidth]{{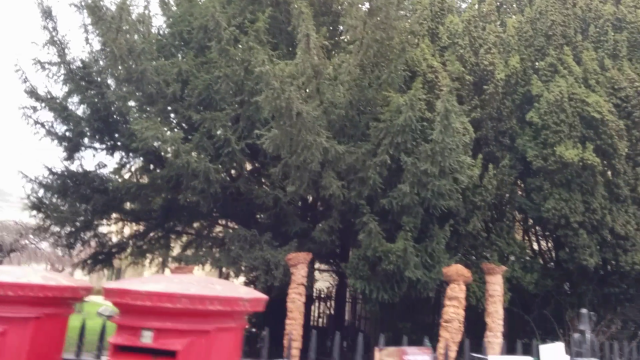}}
  		\caption{}
  		\label{fig:posenet2}
	\end{subfigure}
	\begin{subfigure}{.1125\textwidth}
  		\centering
  		\includegraphics[bb = 0 0 640 360, width=1\textwidth]{{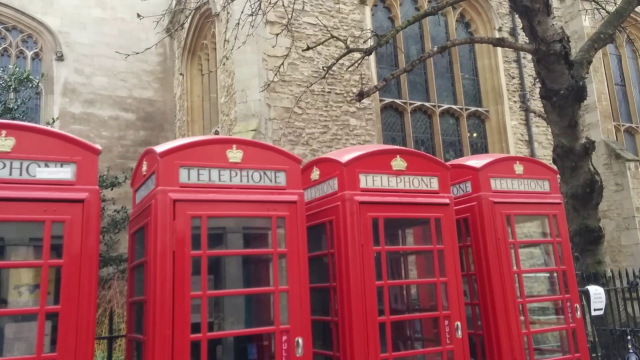}}
  		\caption{}
  		\label{fig:posenet3}
	\end{subfigure}
	\begin{subfigure}{.1125\textwidth}
  		\centering
  		\includegraphics[bb = 0 0 640 360, width=1\textwidth]{{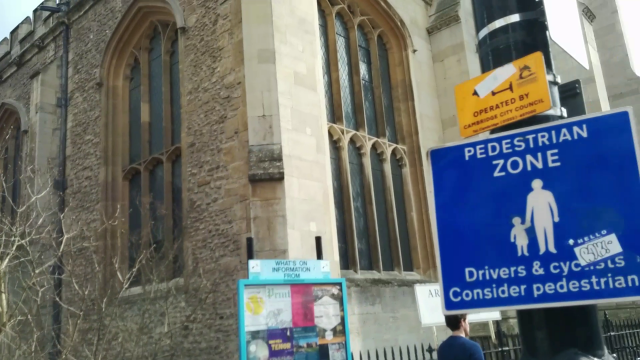}}
  		\caption{}
  		\label{fig:posenet4}
	\end{subfigure}
	\caption{Typical views from the PoseNet dataset \cite{kendall:grimes:2015}.}
	\label{fig:posenet}
\end{figure}

We tested the CNN-F architecture's performance with the St Marys Church sequence, a large scale outdoor scene \cite{kendall:grimes:2015b}, as shown in Figure \ref{fig:posenet}. We obtained a relocalization mean error of 8.54 meters, that is in the same magnitude order as that reported in \cite{kendall:grimes:2015} [2.65 m in a modified GoogLeNet architecture with 23 layers]. 


\begin{figure}[h]
	\begin{center}
	\includegraphics[width=0.45\textwidth]{{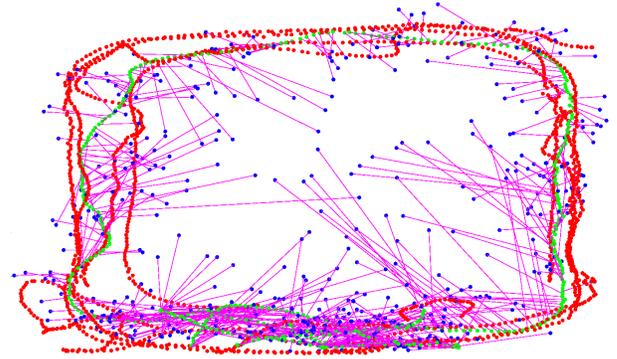}}
	\caption{Relocalisation output applying a CNN-F in the St Marys Church sequence \cite{kendall:grimes:2015b}. The red line is the training sequences and the green one the test sequence. Blue points are the output of the system.}
	\label{fig:posenetrel}
	\end{center}
\end{figure}

In Figure \ref{fig:posenetrel} it can be observed our implementation's output. We use 100 epochs and barely vary the hyperparameters, because at this point the main goal was to design a system that allowed us to perform relocalisation using a CNN. Therefore, we associate the difference in precision compared with that in PoseNet to the mentioned lack of training and the use of a smaller CNN; however, the general performance is maintained.

\section{CNN Map Representation} \label{sect:cnnmap}

From a human observer point of view, it is common to think of metric or topological relationships among elements in space to build maps; for this reason, metric and topological are two well known map representations (such as \cite{cummins:2008}, \cite{angeli:2008}, and \cite{lim:frahm:2012}). However, other less intuitive map representation have been proposed -- e.g. \cite{milford:2012} defines a map as a collection of images and uses image batch matching to find the current position in the map. 

Overall it can be argued that the map representation need not conform to a {\it single} representation type, and that the task and other constraints can lead to different manners in which a map can be represented. Ultimately, for a robotic agent, maps are likely built to be explored or more generally re-explored. It is thus that we highlight once more that relocalisation is a good measure of map effectiveness. In this context the actual map representation used is less relevant as long as it improves relocalisation.

In this work, we represent a map as a regression function $\hat{p} = cnn(I)$ where an element of the population is an input $I$ and its associates output $p$ (e.g. a RGB image and the 6DOF camera pose). The parameters in the regressor $cnn$ are optimized from a population sample (training data); the more representative the sample, the more accurate the model \cite{james:witten:hastie:tibshirani:2014}. 

Finally, we introduce the notion of map compression using CNN map representation. We first define a sample as a collection of elements $(I,p)$ that defines a sensor's traveled trajectory. From a series of samples, we define a regressor $\hat{p} = cnn(I)$ that minimizes the error $|p - \hat{p}|$. This regressor, once defined, is of constant size, and should improve its performance while increasing the population sample (i.e. the number of training trajectories). A compact map representation is then posed as the problem of finding an optimal CNN architecture that keeps minimum relocalisation error values.


\section{EXPERIMENTS AND RESULTS} \label{sect:experiments}

\subsection{CNN for camera relocalisation}

\begin{figure}[h]
	\begin{subfigure}{.1125\textwidth}
		\centering
		\includegraphics[bb = 0 0 640 480, width=1\textwidth]{{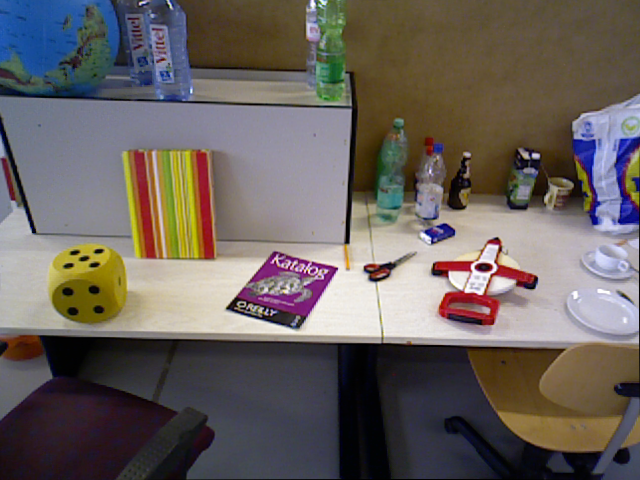}}
		\caption{}
		\label{fig:tum1}
	\end{subfigure}
	\begin{subfigure}{.1125\textwidth}
  		\centering
  		\includegraphics[bb = 0 0 640 480, width=1\textwidth]{{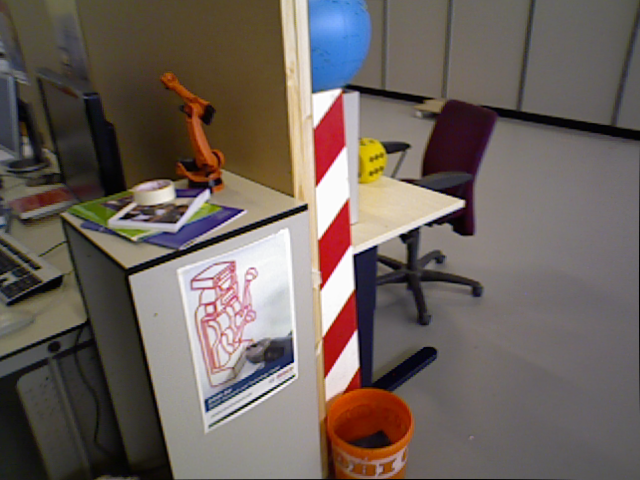}}
  		\caption{}
  		\label{fig:tum2}
	\end{subfigure}
	\begin{subfigure}{.1125\textwidth}
  		\centering
  		\includegraphics[bb = 0 0 640 480, width=1\textwidth]{{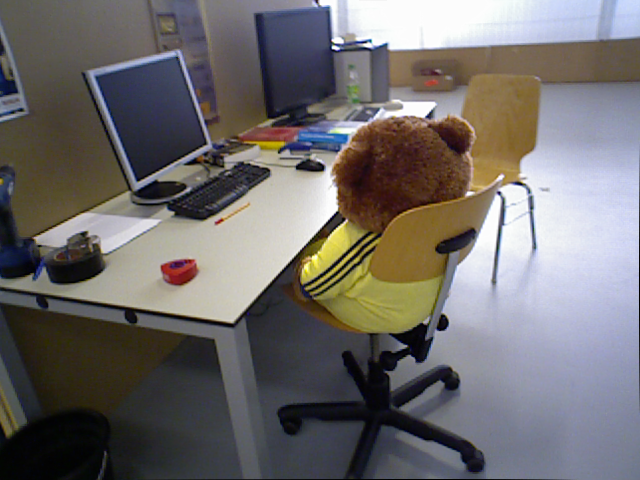}}
  		\caption{}
  		\label{fig:tum3}
	\end{subfigure}
	\begin{subfigure}{.1125\textwidth}
  		\centering
  		\includegraphics[bb = 0 0 640 480, width=1\textwidth]{{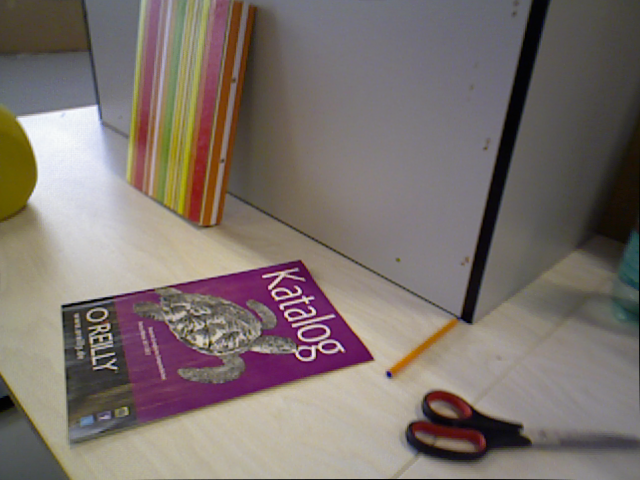}}
  		\caption{}
  		\label{fig:tum4}
	\end{subfigure}
	\caption{Some scenes in the TUM \cite{sturm:2012} long office sequence.}
	\label{fig:tum}
\end{figure}

We first evaluate this implementation with the TUM's long household and office sequence \cite{sturm:2012} -- a texture and structure rich scene, with 21.5m in 87.09s (2585 frames), and a validation data set with 2676 extra frames as can be seen in Figure \ref{fig:tum}.

Learning the external parameters in a CNN is time consuming because there is not really an optimal approach to this task; for this reason, the use of generic image representations has been proposed such as  \cite{razavian:2014}, where the models trained in one task can be used in another.

\begin{figure}[h]
	\begin{subfigure}{.15\textwidth}
		\centering
		\includegraphics[width=1\textwidth]{{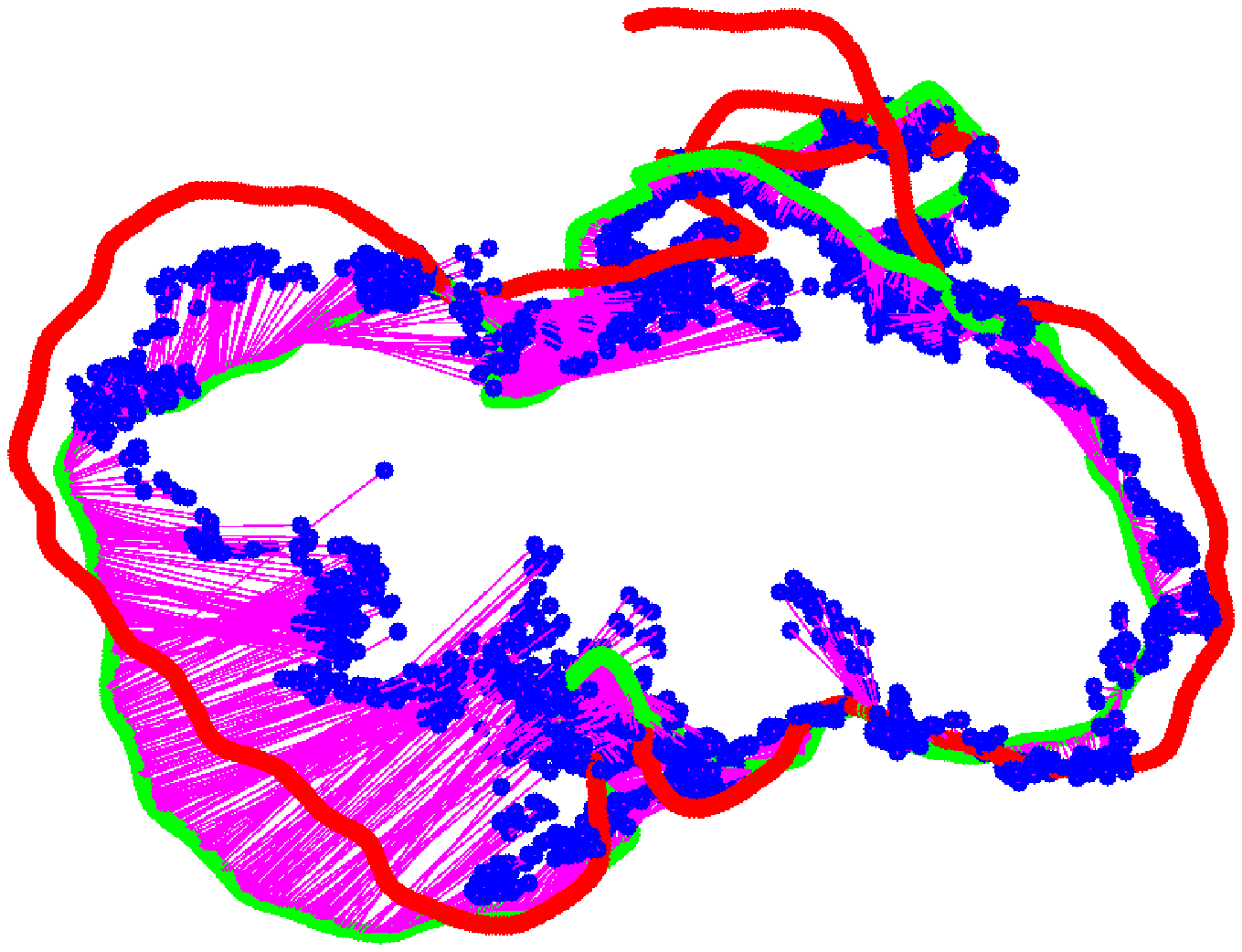}}
		\caption{}
		\label{fig:tumrel1}
	\end{subfigure}
	\begin{subfigure}{.15\textwidth}
  		\centering
  		\includegraphics[width=1\textwidth]{{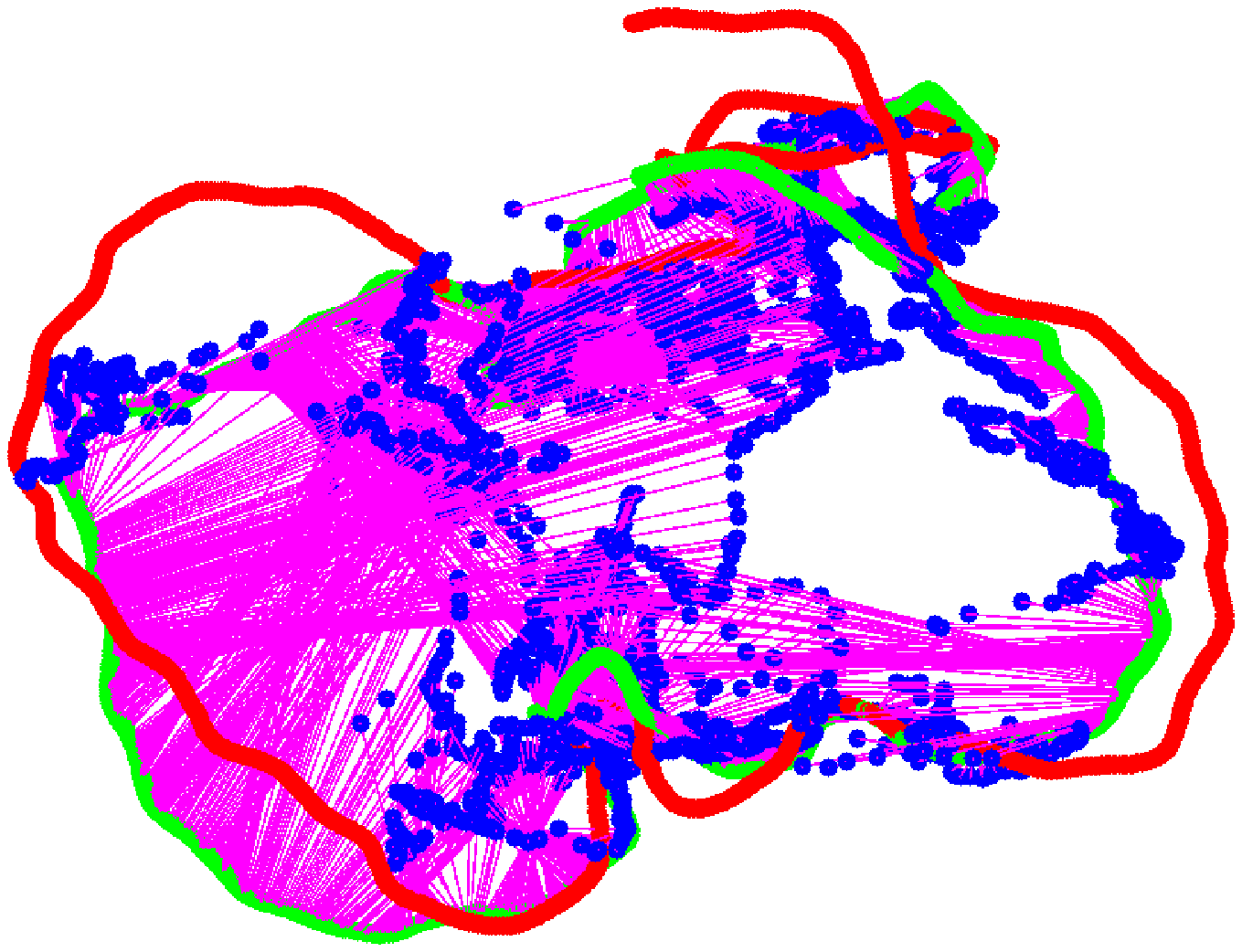}}
  		\caption{}
  		\label{fig:tumrel2}
	\end{subfigure}
	\begin{subfigure}{.15\textwidth}
  		\centering
  		\includegraphics[width=1\textwidth]{{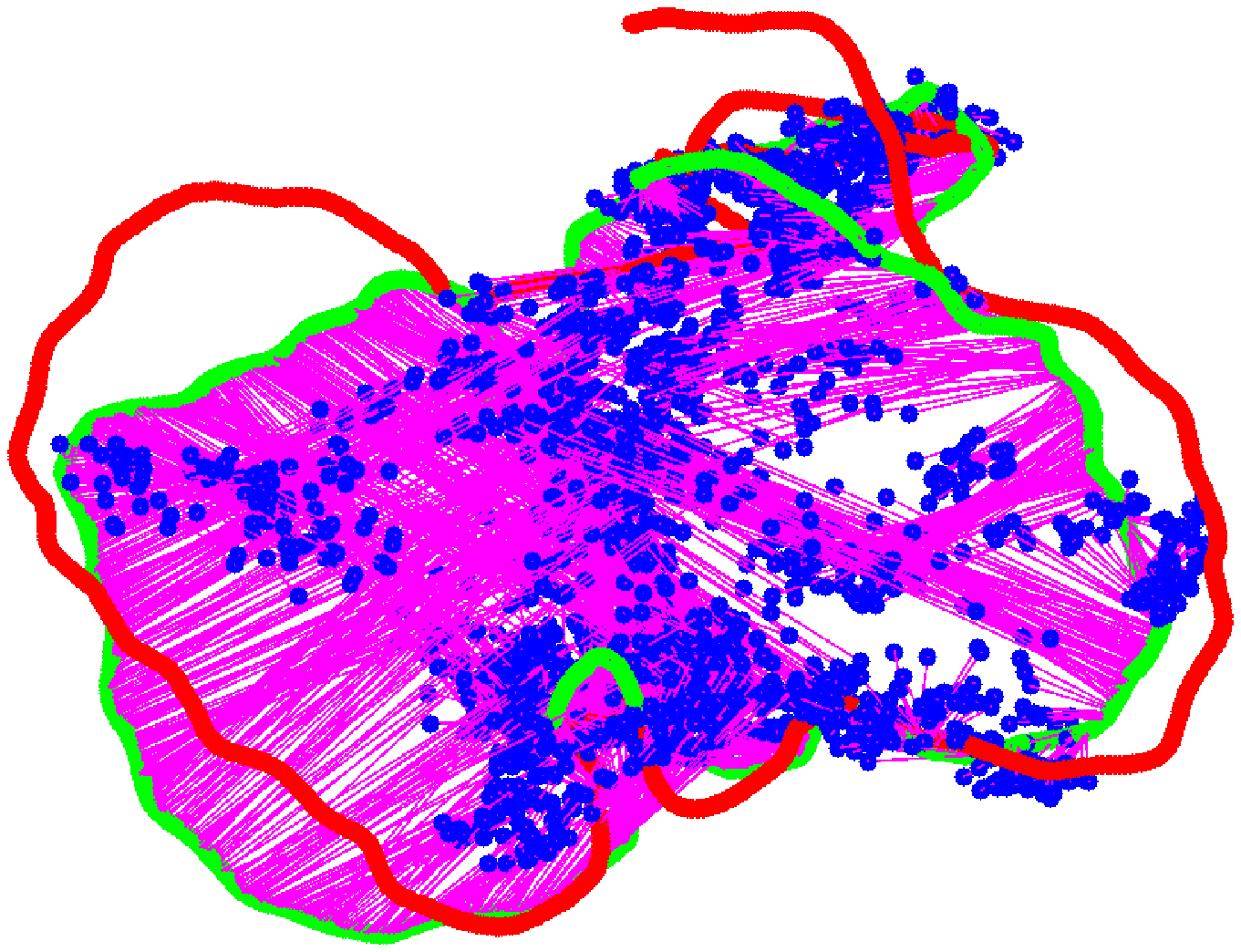}}
  		\caption{}
  		\label{fig:tumrel3}
	\end{subfigure}
	\caption{Relocalisation performance in the TUM sequence \cite{sturm:2012}: a) RGB input with a pre-trained CNN, b) RGB input and a randomly initialized CNN, and c) 3D point cloud input. Red line indicates the training trajectory while the green line is the testing one. Blue points are the output of the neural network.}
	\label{fig:tumrel}
\end{figure}

We tested the CNN-F with pre-trained wights optimized for object detection in the ImageNet dataset \cite{deng:dong:2009}, as well as with random weights. The response to a RGB input can be seen in Figure \ref{fig:tumrel} -- where the red line indicates the training sequence and the green one the test sequence -- with a relocalisation mean error of 0.572 meters using pre-trained weights, and 0.867 meters when random weights were used.

\begin{figure}[ht]
	\center
	\begin{subfigure}{.225\textwidth}
		\centering
		\includegraphics[width=1\textwidth]{{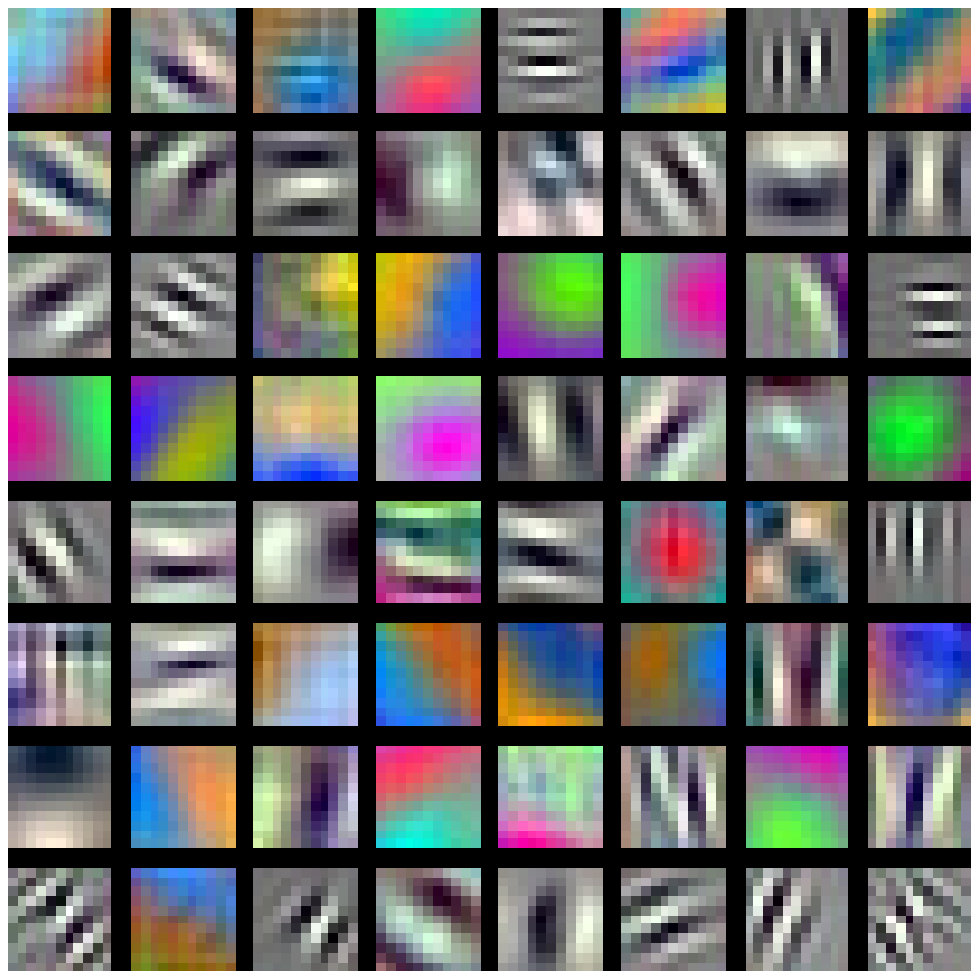}}
		\caption{}
		\label{fig:filters0}
	\end{subfigure}
	\begin{subfigure}{.225\textwidth}
  		\centering
  		\includegraphics[width=1\textwidth]{{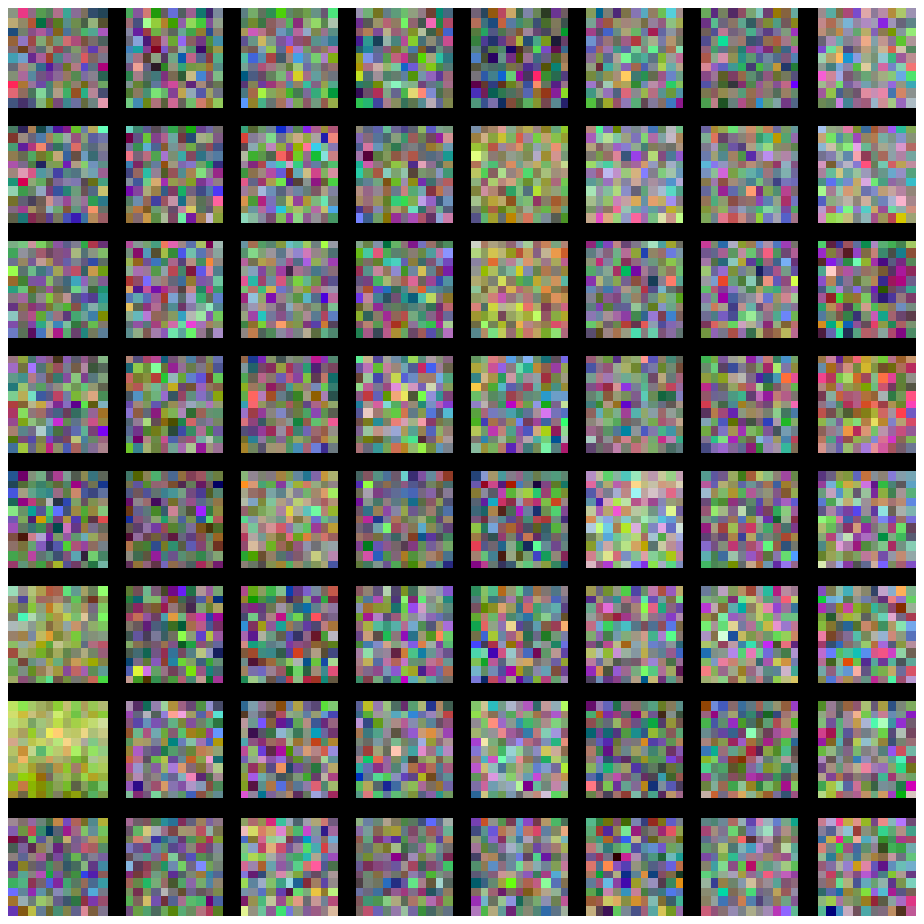}}
  		\caption{}
  		\label{fig:filters1}
	\end{subfigure}
	\caption{First layer filters visualisation for a) pre-trained weights after 250 finetunning training epochs and b) randomly initialized filters after training.}
	\label{fig:filters}
\end{figure}

We associate this operation difference to the pre-training process. Figure \ref{fig:filters} visualize the filters after 250 epochs; more structured filters -- even though they were trained for a different task -- showed better relocalisation performance. 

Regarding the other sensor inputs, as pre-trained CNN is more common for RGB than depth or spatial  data, in this experiment we tested randomly initialized CNNs to make a fairer comparison among them. In Table \ref{tab:cnntum} we can see the relocalisation mean error for all inputs, where RGB information seems to outperform the other.

\begin{table}[!h]
	\caption{CNN-F relocalisation mean error [in meters] for different inputs after 1000 epochs in the long office sequence from the TUM dataset \cite{sturm:2012}.}
	\label{tab:cnntum}
	\begin{center}
	\resizebox{0.45\textwidth}{!}{
		\begin{tabular}{c|c}
			&Relocalisation mean error [m]\\ 
		\hline
			Depth & 1.926 $\pm$ 0.595\\
			Gray & 0.936 $\pm$ 0.814\\
            Point Cloud & 1.191 $\pm$ 0.960\\
            RGB & 0.877 $\pm$ 0.820\\
            \textbf{Pre-trained RGB} & \textbf{0.572 $\pm$ 0.492}\\
            RGB+Depth & 1.002 $\pm$ 0.817\\
            RGB+Point Cloud & 2.542 $\pm$ 0.456\\
		\hline
		\end{tabular}}
	\end{center}
\end{table}

We noticed that, when combine information layers are used, the performance decreases. It might be due to the difference in the input nature (color, depth, and spatial position). One way to solve it can be using parallel networks and then average the output, as in \cite{agrawal:2015}.

\begin{figure*}[ht]
	\center
	\begin{subfigure}{.225\textwidth}
		\centering
		\includegraphics[width=1\textwidth]{{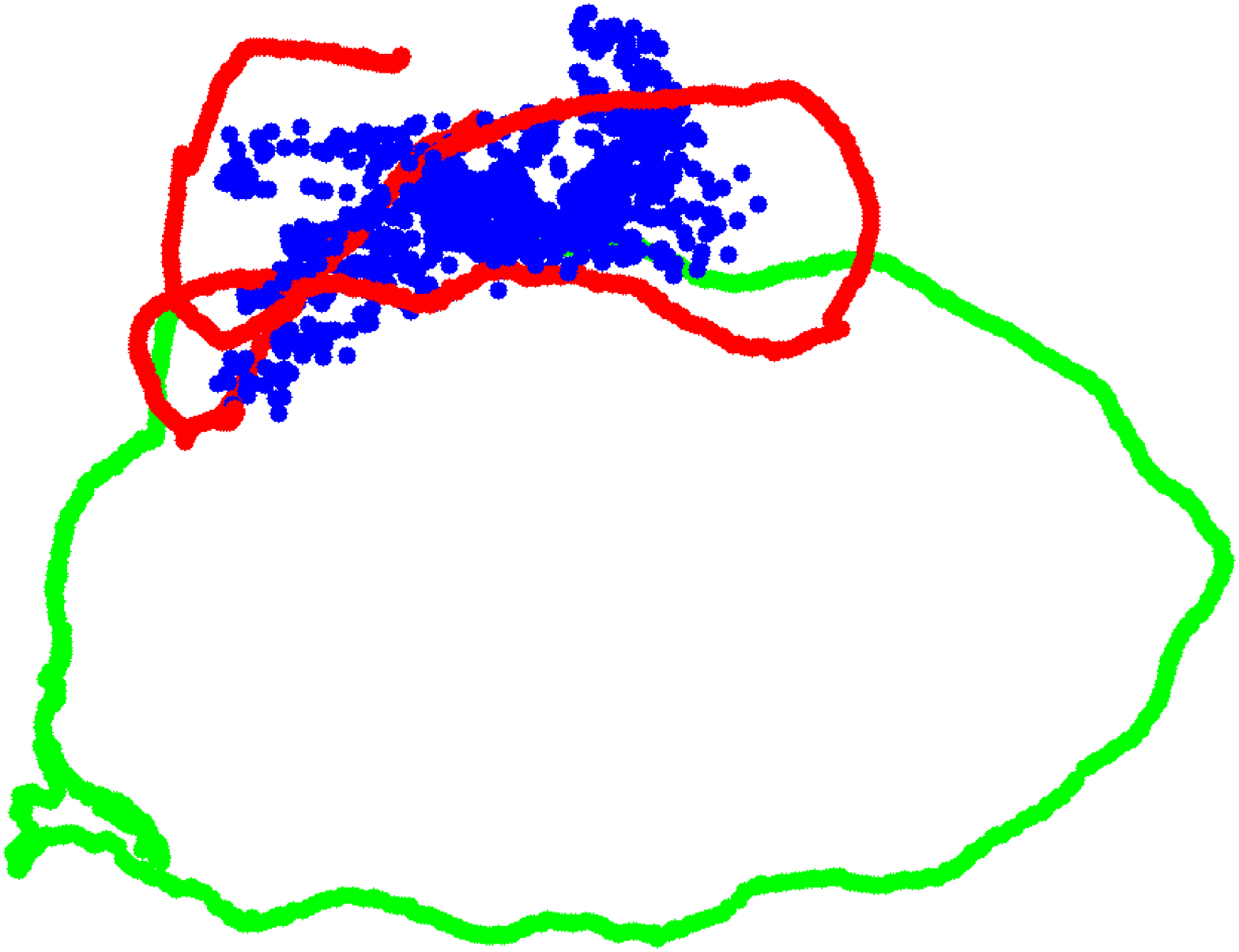}}
		\caption{}
		\label{fig:redkitchen0}
	\end{subfigure}
	\begin{subfigure}{.225\textwidth}
  		\centering
  		\includegraphics[width=1\textwidth]{{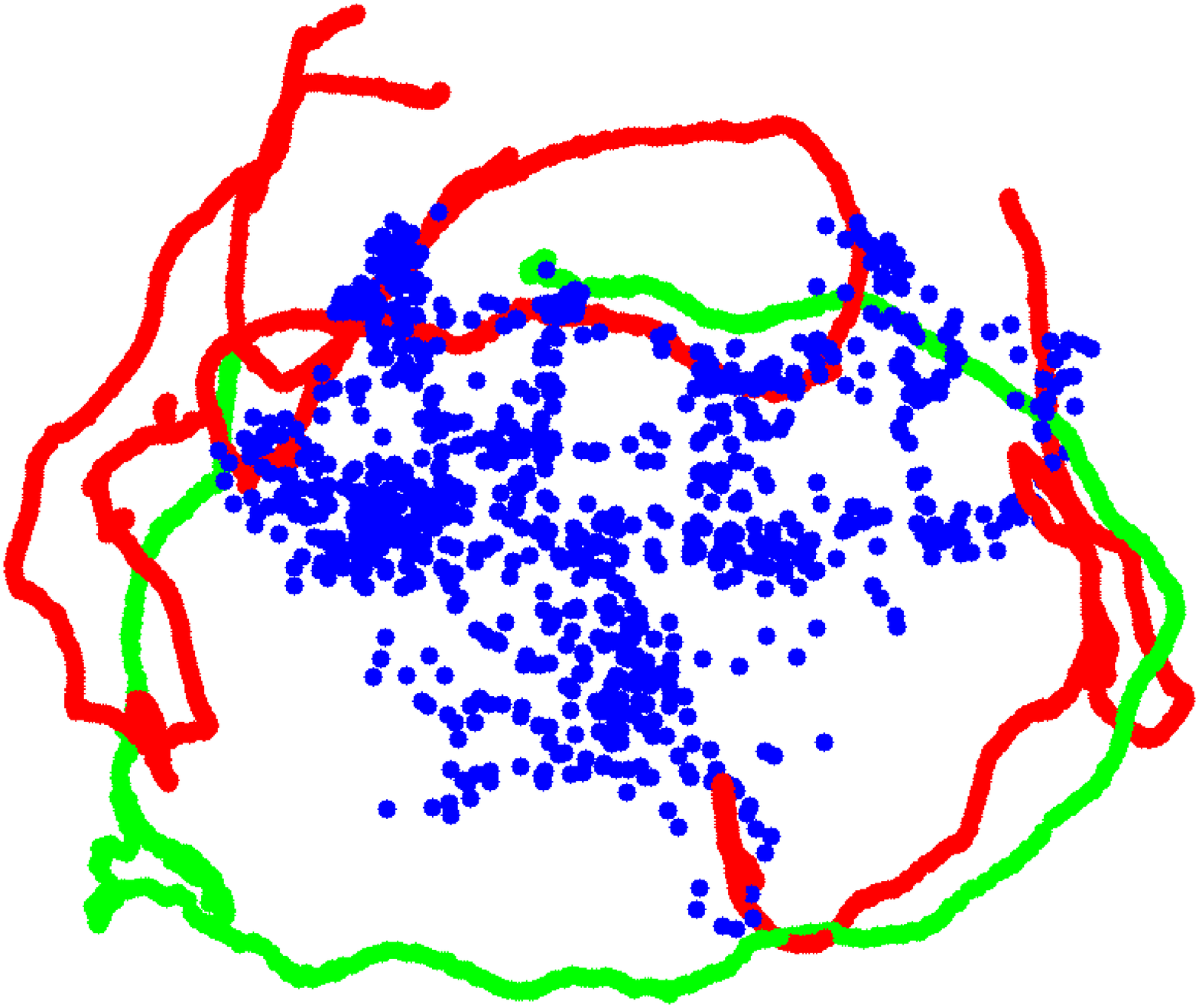}}
  		\caption{}
  		\label{fig:redkitchen1}
	\end{subfigure}
	\begin{subfigure}{.225\textwidth}
  		\centering
  		\includegraphics[width=1\textwidth]{{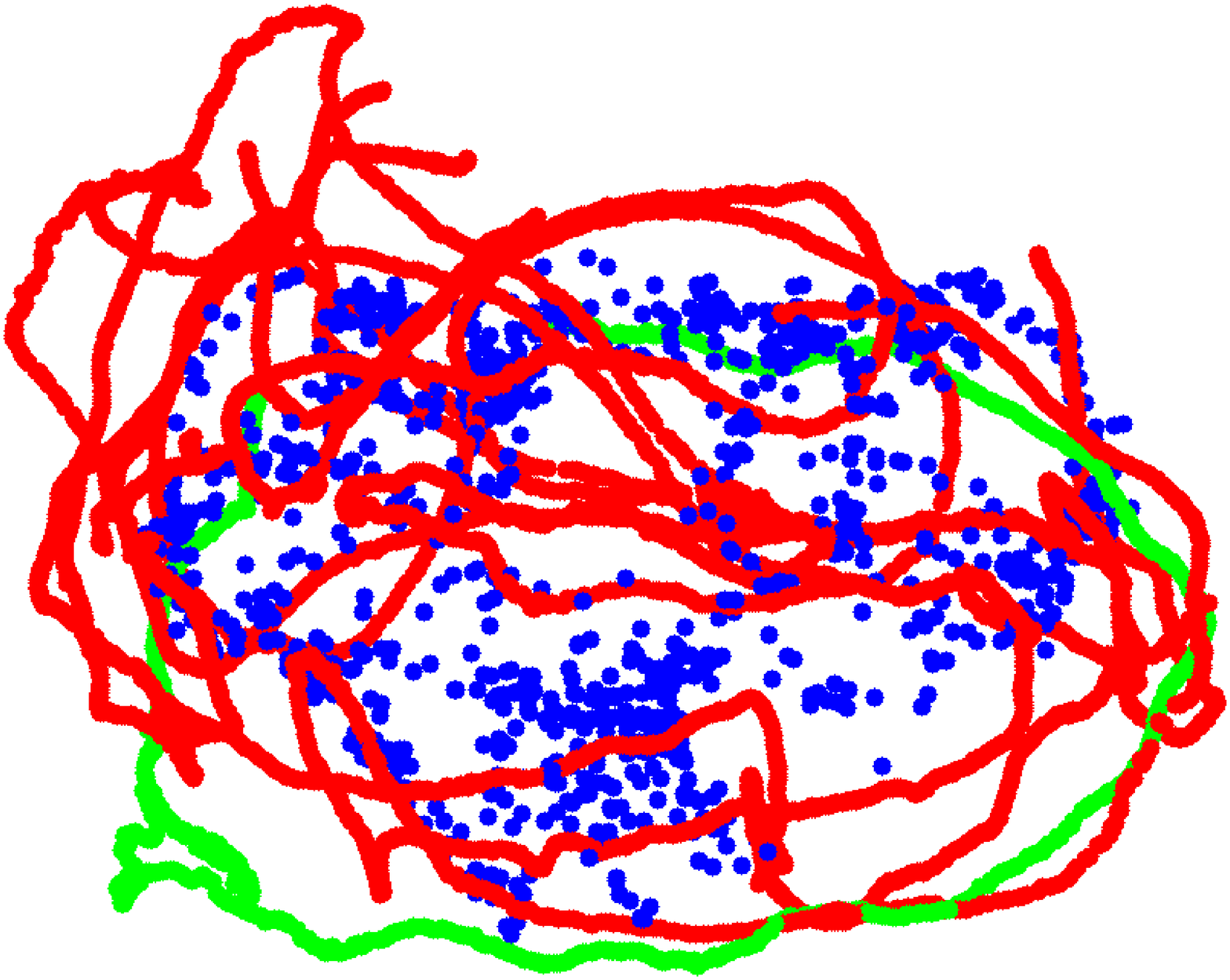}}
  		\caption{}
  		\label{fig:redkitchen2}
	\end{subfigure}
	\begin{subfigure}{.225\textwidth}
  		\centering
  		\includegraphics[width=1\textwidth]{{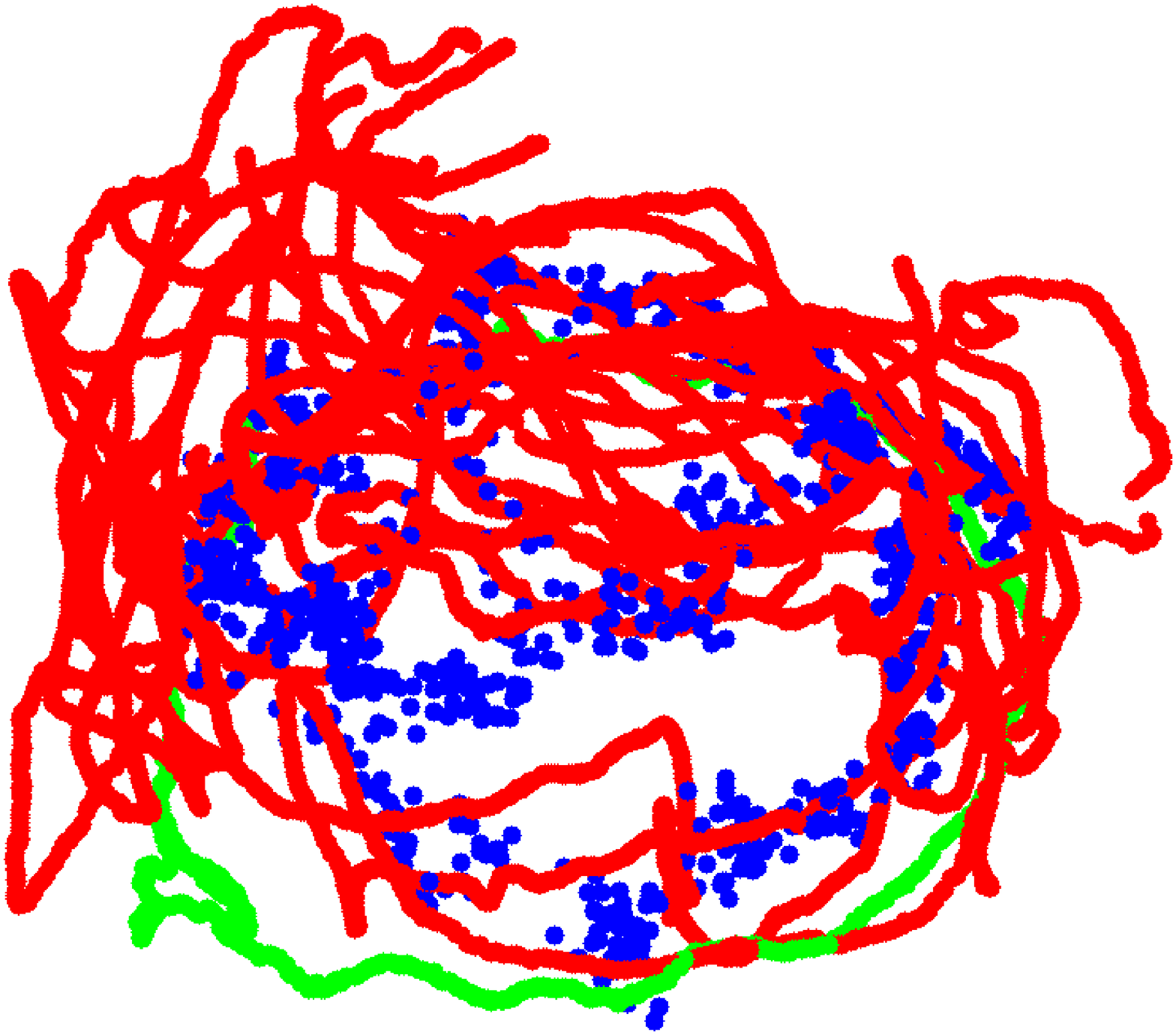}}
  		\caption{}
  		\label{fig:redkitchen3}
	\end{subfigure}
	\caption{Relocalisation performance in from the Red Kitchen sequence in 7-Scenes dataset \cite{glocker:izadi:2013} after training with a) one, b) three, c) seven, and d) eleven sequences. Red lines indicate the training sequences and the green line is the test one; blue points are the output of the system.}
	\label{fig:redkitchen}
\end{figure*}

\begin{figure*}[ht]
	\center
	\begin{subfigure}{.225\textwidth}
		\centering
		\includegraphics[width=1\textwidth]{{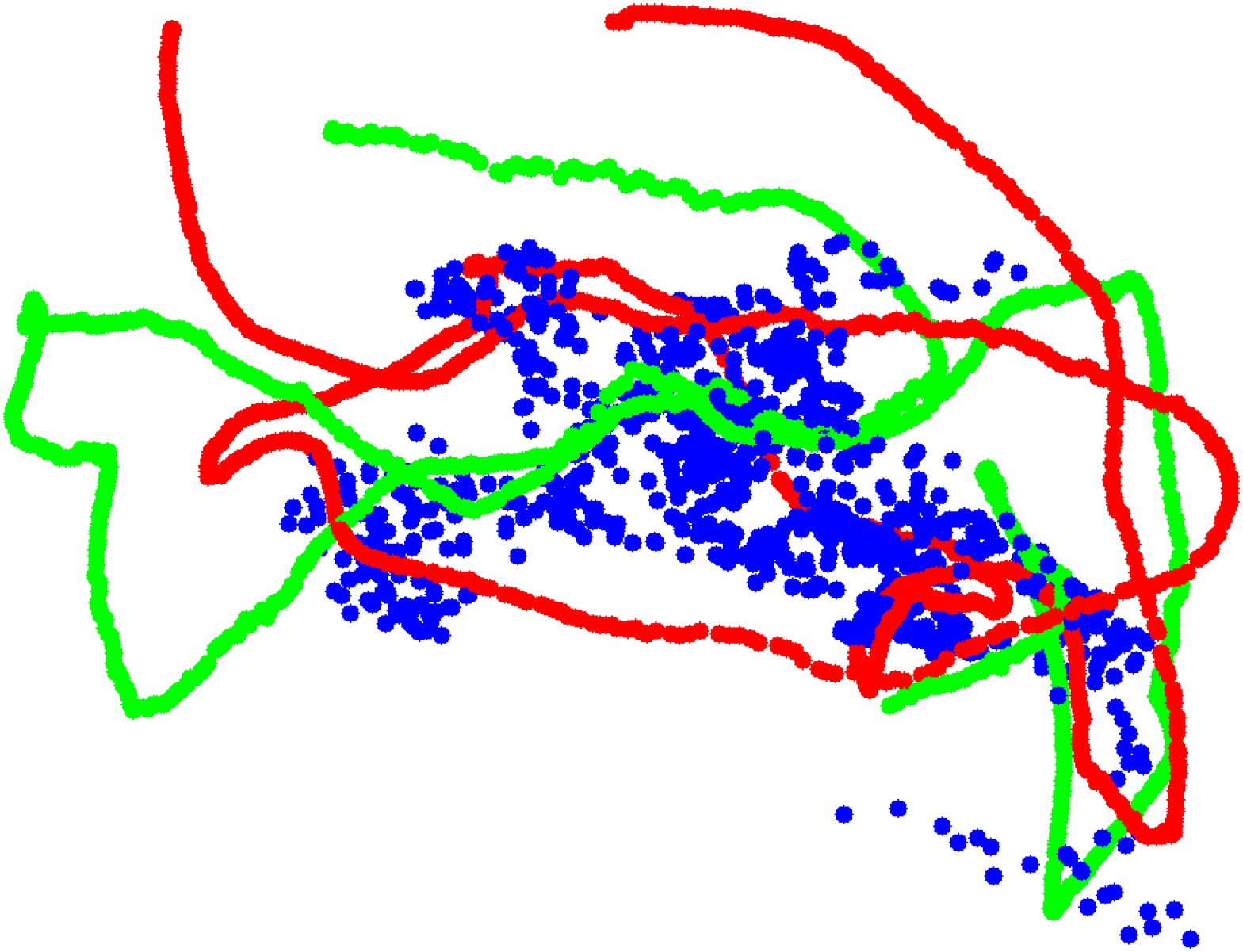}}
		\caption{}
		\label{fig:office0}
	\end{subfigure}
	\begin{subfigure}{.225\textwidth}
  		\centering
  		\includegraphics[width=1\textwidth]{{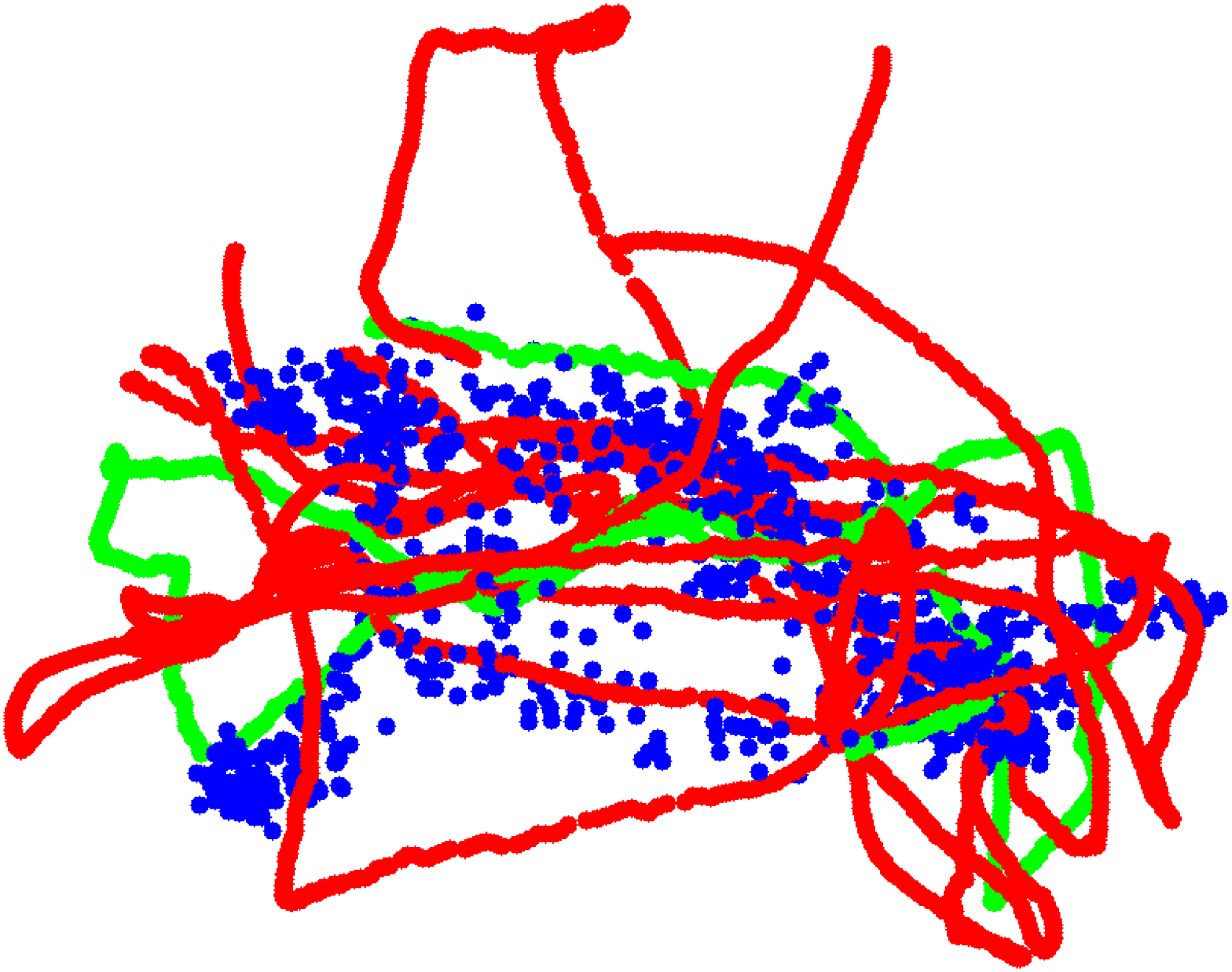}}
  		\caption{}
  		\label{fig:office1}
	\end{subfigure}
	\begin{subfigure}{.225\textwidth}
  		\centering
  		\includegraphics[width=1\textwidth]{{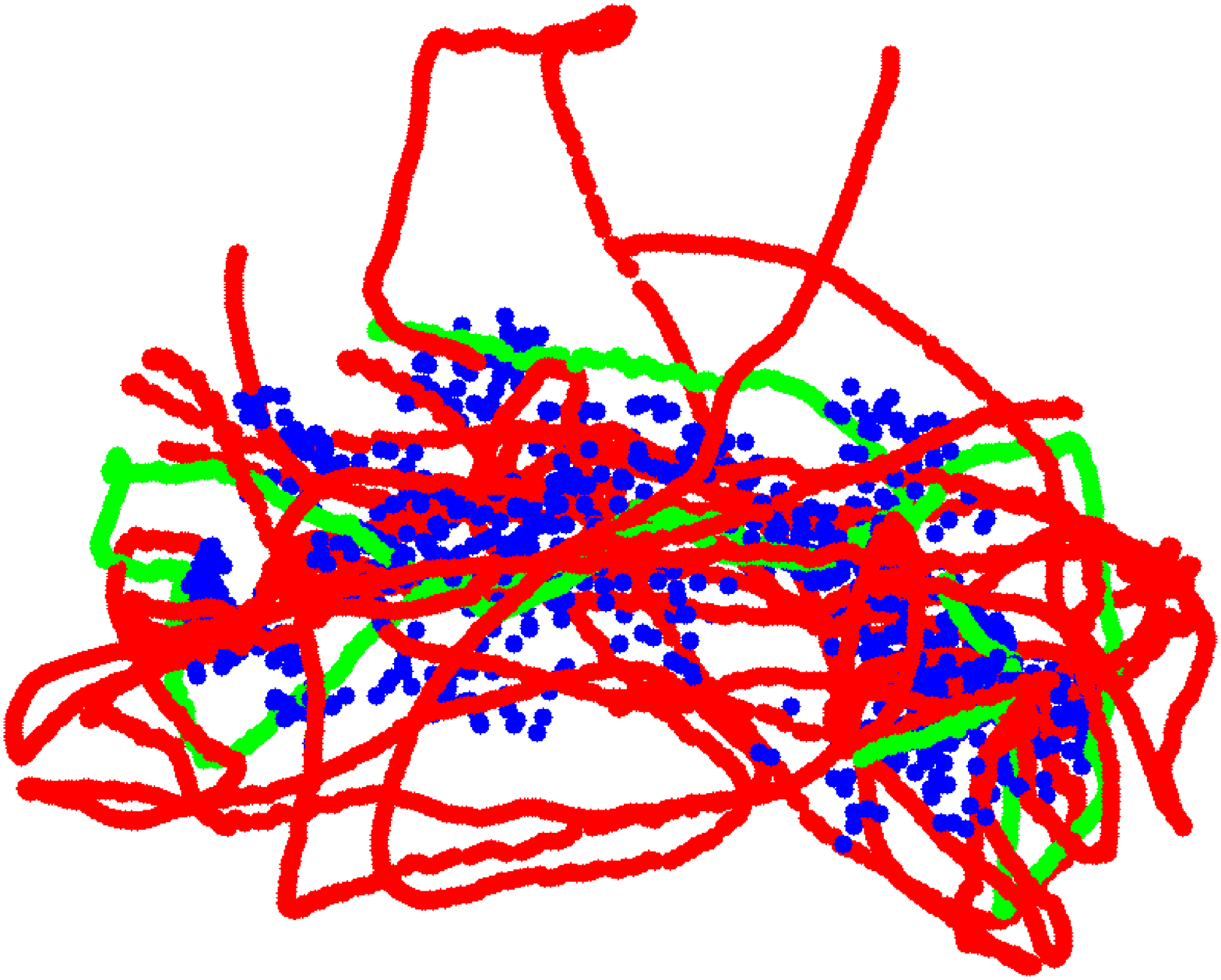}}
  		\caption{}
  		\label{fig:office2}
	\end{subfigure}
	\begin{subfigure}{.225\textwidth}
  		\centering
  		\includegraphics[width=1\textwidth]{{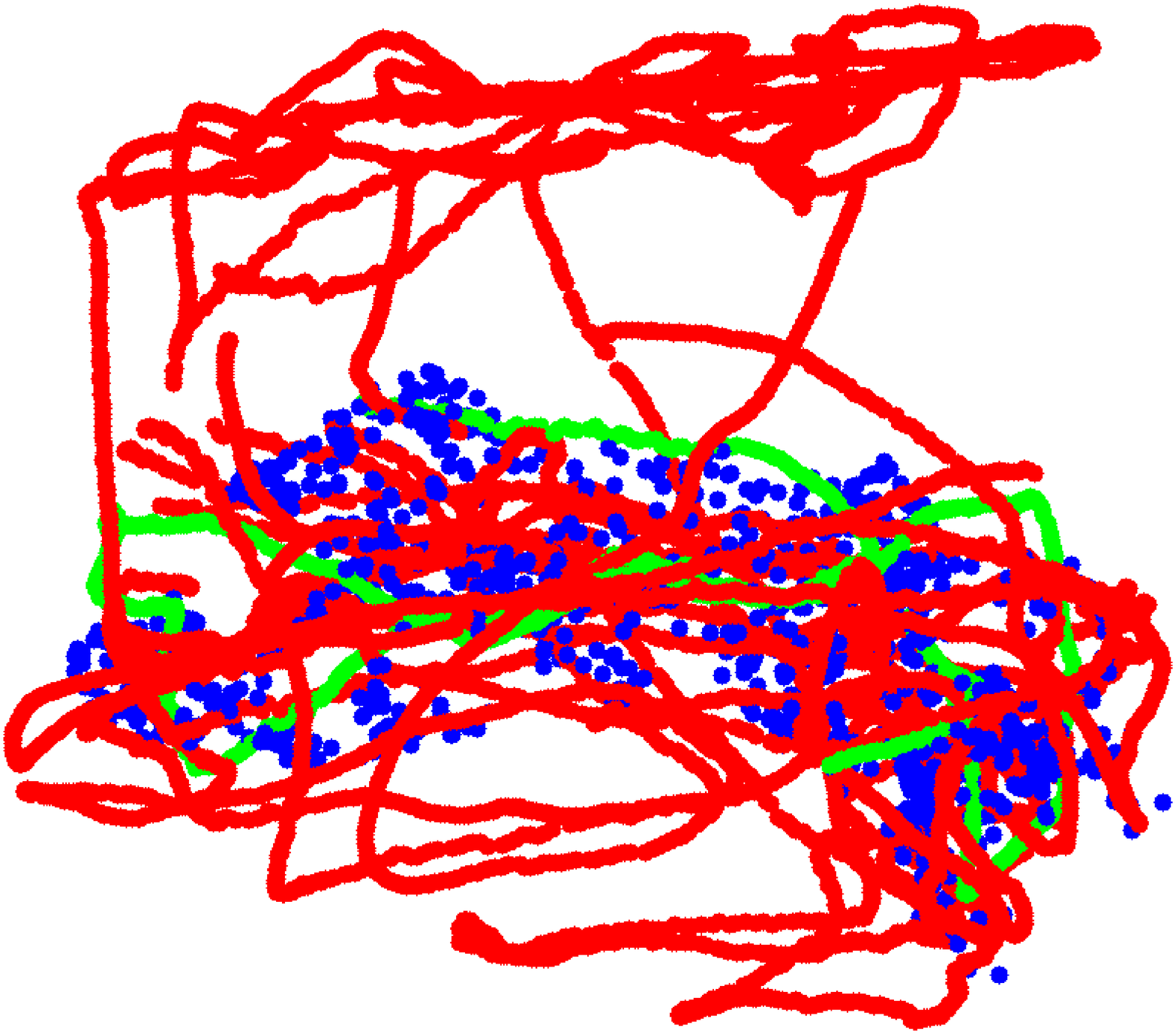}}
  		\caption{}
  		\label{fig:office3}
	\end{subfigure}
	\caption{Relocalisation performance in the 7-Scenes' Office sequence \cite{glocker:izadi:2013} after training with a) one, b) four, c) six, and d) nine sequences. Colors are as in Figure \ref{fig:redkitchen}.}
	\label{fig:office}
\end{figure*}

\begin{figure}[h]
	\begin{center}
	\includegraphics[width=0.45\textwidth]{{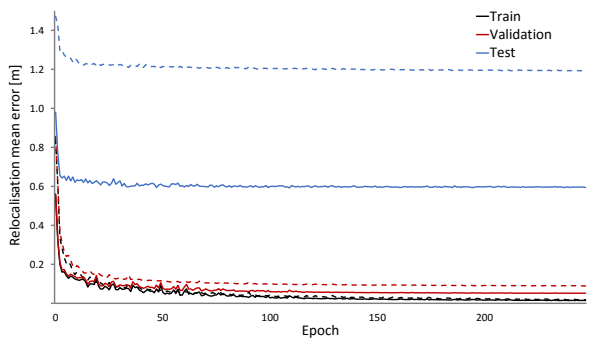}}
	\caption{Relocalisation mean error at different inputs. A continuous line corresponds to an RGB input while a dotted line corresponds to a 3D position point cloud input in the TUM dataset \cite{sturm:2012}. We observe a similar training behavior, but the test performance diverges.}
	\label{fig:rgbvspc}
	\end{center}
\end{figure}

Overfitting -- i.e., a parameter saturation in simple model -- also affects the performance. In Figure \ref{fig:rgbvspc} we observe that even though the RGB and point cloud training and validation processes behave similarly, in the test set the differences are notorious. To overcome this issue, we can use a more complex model or pre-trained filters in bigger datasets -- the use of 3D primitives to retrieve information useful for image understanding was explored in \cite{fouhey:2013}, while \cite{jimenez:2016} learned generative models of 3D structures from volumetric data.

\subsection{Multiple trajectories learning}

\begin{figure}[h]
	\begin{subfigure}{.1125\textwidth}
		\centering
		\includegraphics[bb = 0 0 640 480, width=1\textwidth]{{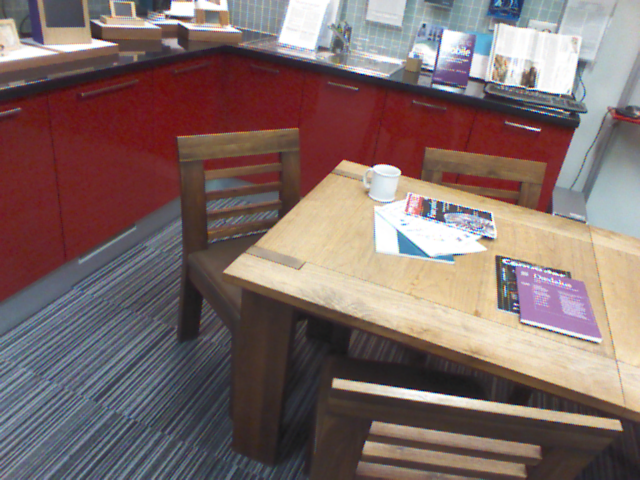}}
		\caption{}
		\label{fig:7scenes1}
	\end{subfigure}
	\begin{subfigure}{.1125\textwidth}
  		\centering
  		\includegraphics[bb = 0 0 640 480, width=1\textwidth]{{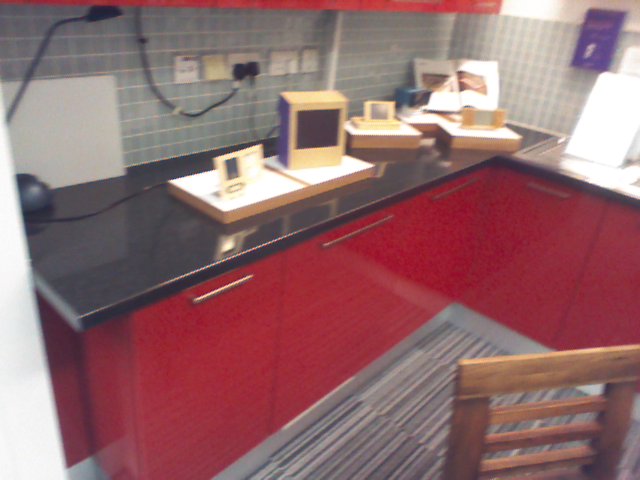}}
  		\caption{}
  		\label{fig:7scenes2}
	\end{subfigure}
	\begin{subfigure}{.1125\textwidth}
  		\centering
  		\includegraphics[bb = 0 0 640 480, width=1\textwidth]{{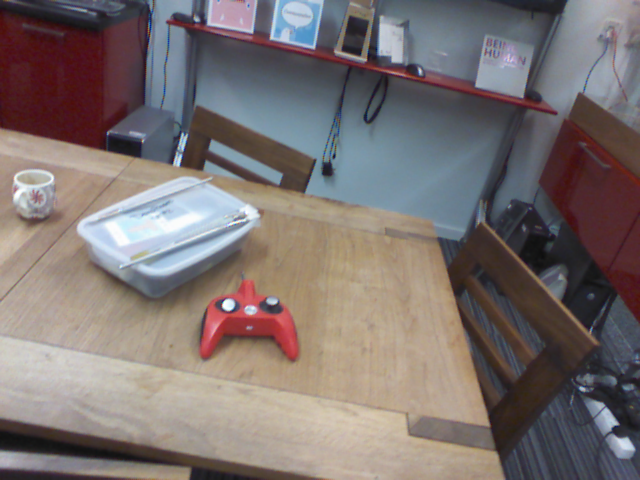}}
  		\caption{}
  		\label{fig:7scenes3}
	\end{subfigure}
	\begin{subfigure}{.1125\textwidth}
  		\centering
  		\includegraphics[bb = 0 0 640 480, width=1\textwidth]{{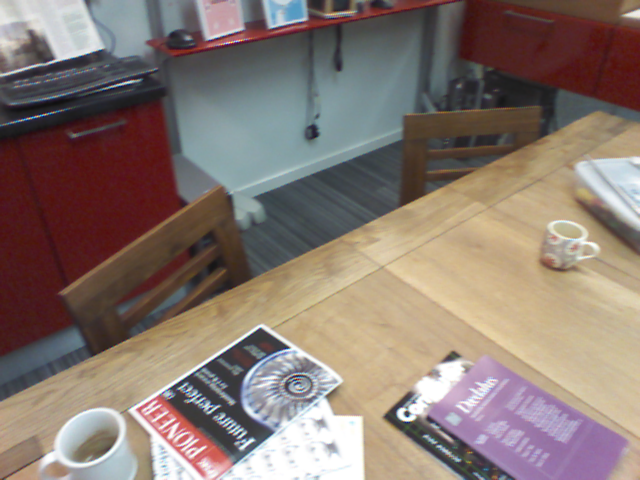}}
  		\caption{}
  		\label{fig:7scenes4}
	\end{subfigure}
	\caption{Typical views from the Red Kitchen sequence in the 7-Scenes dataset \cite{glocker:izadi:2013}.}
	\label{fig:7scenes}
\end{figure}

To test the relocalisation performance with respect to the number of training sequences, and hence its capability for compression, we use the 7-Scenes dataset (\cite{glocker:izadi:2013}, \cite{shotton:glocker:2013}), as shown in Figure \ref{fig:7scenes}; this dataset consists of several trajectories taken by different persons moving around the same environment. Training, validation and testing sequences are indicated in the dataset itself.

We first tested this implementation with RGB data and compared it with PoseNet results, and then we evaluate this architecture for all data inputs; results are shown in Table \ref{tab:cnn7scenes}. Although RGB error is again the lowest, with more training images per set, similar performances are found between different data types. However, for the rest of the section we will use only RGB data.

In this test, using RGB data as the sensor input, we observe that CNN-F (8 layers) behaves at least as good as PoseNet, a 23 layers and more complex CNN, where both of them have a relocalisation mean error of 0.59 meters. It remains an open problem the task of designing customized CNN map representation by systematically modifying the neural network architecture itself.

\begin{table}[!h]
	\caption{CNN-F relocalisation mean error [in meters] for different inputs after 100 epochs in the Red Kitchen sequence from the 7-Scenes dataset\cite{glocker:izadi:2013}. In italics we present the PoseNet mean error reported in \cite{kendall:grimes:2015}.}
	\label{tab:cnn7scenes}
	\begin{center}
	\resizebox{0.45\textwidth}{!}{
		\begin{tabular}{c|c}
			Input&Relocalisation mean error [m]\\ 
		\hline
			Depth & 1.326 $\pm$ 0.372\\
			Gray & 0.795 $\pm$ 0.504\\
            Point Cloud & 0.791 $\pm$ 0.692\\
            \textbf{RGB} & \textbf{0.589 $\pm$ 0.425}\\
            \textit{PoseNet (RGB)} & \textit{0.59}\\
            RGB+Depth & 0.748 $\pm$ 0.551\\
            RGB+Point Cloud & 0.685 $\pm$ 0.682\\
		\hline
		\end{tabular}}
	\end{center}
\end{table}

Furthermore, in traditional mapping techniques, in general the map increases when new views are added; instead, by using a CNN map representation, we increasing the map information while maintaining a constant neural network size by re-training it when new trajectories are added.

\begin{figure}[ht]
	\begin{center}
	\includegraphics[width=0.45\textwidth]{{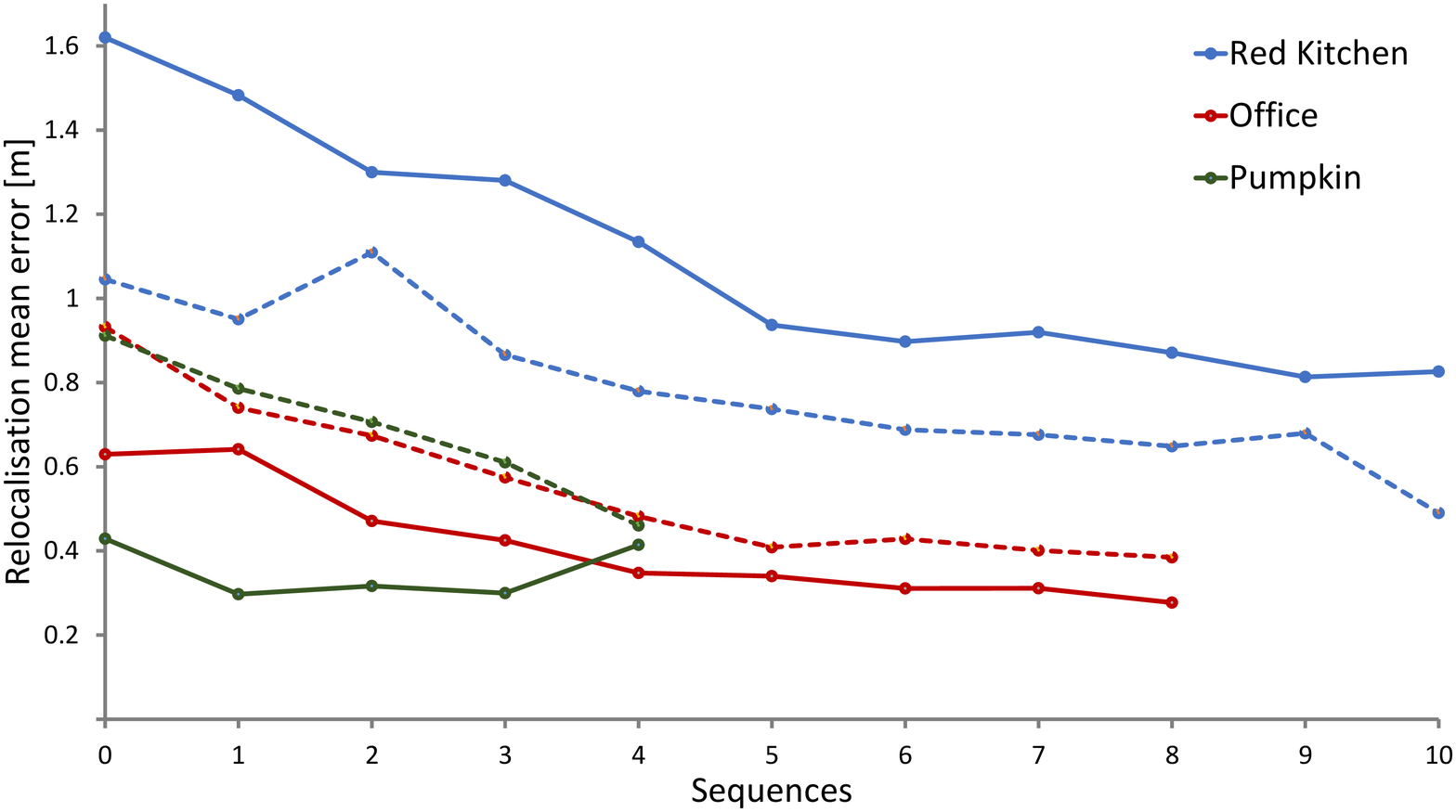}}
	\caption{Relocalisation mean error in the 7-Scenes dataset \cite{glocker:izadi:2013} with respect to the number of training trajectories (we fix one sequence for testing and use the remaining for training; dotted lines indicates a different test sequence). While the number of training trajectories increases, the error decreases but the CNN-F size remains the same (we only affect the weights).}
	\label{fig:7scenescumul}
	\end{center}
\end{figure}

We used the Red Kitchen, Office and Pumpkin sequences in the 7-Scenes dataset. We took out one of the trajectories for testing and gradually train the CNN adding one remaining sequence at a time. Figure \ref{fig:7scenescumul} shows that while increasing the number of trajectories, precision also increases but, by construction, the size of the CNN remains the same, as expected. Nevertheless, we didn't reach an asymptotic behavior that would have indicated that the neural network was saturated; this indicates that more trajectories can still be added, improving the performance. While compact, then, this map representation is also constant-size when new information is added.

In Figure \ref{fig:redkitchen} we can see some outputs for the Red Kitchen sequence where the relocalisation improves as we add more trajectories; in Figure \ref{fig:office} we observe the same behavior in the Office sequence. There, we observe how the relocalised cameras (blue points) are closer to the test sequence (green line) when more training sequences are present (red line).

\section{CONCLUSIONS}

We presented a first approach toward CNN map compression for camera relocalisation. We studied the response to different inputs and to different trajectories. We first show that for these kind of models, the RGB images present the best performance, compared with other types of data, as depth or 3D point clouds. We introduced the idea of map compression as the task of finding optimal CNN architectures -- we obtain similar performance using CNN-F (8 layers) than PoseNet (23 layers). Then, we demonstrate that increasing the training trajectories, accuracy increases as well, without increasing the map size -- i.e. only the filters' weights change but not the number of layers.

For future work we note that more complex relocalisation such as semantic or topological relocalisation were not explored here. One potential direction encouraged by these results is to train simpler networks for object recognition with labeled datasets, and use a second network that accepts semantic information as input for relocalisation. This kind of multi-network systems, where two complex systems interact to perform a single task are of interest to expand on the current work.

\addtolength{\textheight}{-15cm} 









\bibliographystyle{ieeetr}
\bibliography{all}

\end{document}